\documentclass[3p,times,procedia]{elsarticle}
\usepackage{ecrc}
\usepackage[bookmarks=false]{hyperref}
    \hypersetup{colorlinks,
      linkcolor=blue,
      citecolor=blue,
      urlcolor=blue}

\volume{00}

\firstpage{1}

\journalname{Procedia Computer Science}

\runauth{Author name}

\jid{procs}

\usepackage{amssymb}

\usepackage[figuresright]{rotating}

\begin{document}

\begin{frontmatter}



\dochead{Proceedings of International Conference on Biomimetic, Intelligence and Robots}

\title{Learning to Adapt Foundation Model DINOv2 for Capsule Endoscopy Diagnosis}


\author[a]{Bowen Zhang\corref{cor1}} 
\author[b]{Ying Chen\corref{cor1}} 
\author[a]{Long Bai}
\author[c]{Yan Zhao}
\author[d]{Yuxiang Sun}
\author[a]{Yixuan Yuan}
\author[c]{Jianhua Zhang}
\author[a,b,e]{Hongliang Ren\corref{cor2}}

\address[a]{Department of Electronic Engineering, The Chinese University of Hong Kong, Hong Kong SAR, China}
\address[b]{Shenzhen Research Institute, The Chinese University of Hong Kong, Shenzhen, China}
\address[c]{School of Mechanical Engineering, University of Science and Technology Beijing, Beijing, China}
\address[d]{Department of Mechanical Engineering, City University of Hong Kong, Hong Kong SAR, China}
\address[e]{Department of Biomedical Engineering, National University of Singapore, Singapore}
\begin{abstract}
Foundation models have become prominent in computer vision, achieving notable success in various tasks. However, their effectiveness largely depends on pre-training with extensive datasets. Applying foundation models directly to small datasets of capsule endoscopy images from scratch is challenging. Pre-training on broad, general vision datasets is crucial for successfully fine-tuning our model for specific tasks. In this work, we introduce a simplified approach called Adapt foundation models with a low-rank adaptation (LoRA) technique for easier customization.
Our method, inspired by the DINOv2 foundation model, applies low-rank adaptation learning to tailor foundation models for capsule endoscopy diagnosis effectively. Unlike traditional fine-tuning methods, our strategy includes LoRA layers designed to absorb specific surgical domain knowledge. During the training process, we keep the main model (the backbone encoder) fixed and focus on optimizing the LoRA layers and the disease classification component. We tested our method on two publicly available datasets for capsule endoscopy disease classification. The results were impressive, with our model achieving 97.75\% accuracy on the Kvasir-Capsule dataset and 98.81\% on the Kvasir-v2 dataset. Our solution demonstrates that foundation models can be adeptly adapted for capsule endoscopy diagnosis, highlighting that mere reliance on straightforward fine-tuning or pre-trained models from general computer vision tasks is inadequate for such specific applications.

\end{abstract}

\cortext[cor1]{These authors contributed to the work equally and should be regarded as co-first authors.}
\cortext[cor2]{Corresponding author.}
\end{frontmatter}

\email{hlren@ee.cuhk.edu.hk}

\section{INTRODUCTION}
\label{main}
The small intestine presents various types of lesions, including ulcers, erosion, vascular dysplasia, and lymphoma, among others ~\cite{tan2024endoood, wang2023rethinking}. The complex and overlapping structures of the small intestine, coupled with their lack of fixed shape, pose challenges for traditional endoscopes to navigate effectively. The emergence of capsule endoscopy ~\cite{zhang2022deep, li2023semi} offers new prospects for small intestine examination. This allows for the acquisition of direct and clear video signals that enable doctors to examine and determine the disease, as well as formulate appropriate treatment plans. Nevertheless, the analysis of capsule endoscopy videos frame-by-frame can be very laborious. Examiners must possess extensive knowledge and experience in small bowel diseases. These challenges contribute to a high rate of misdiagnoses when doctors manually process video information. 

In recent years, deep learning (DL) has demonstrated significant potential in automated diagnosis and analysis for capsule endoscopy diagnosis~\cite{bai2023llcaps, bai2022transformer, he2016deep, touvron2021training}. Real-time artificial intelligence (AI) algorithms analyze images, assisting doctors with rapid diagnoses. This approach has yielded promising results, as good deep-learning algorithms can achieve high diagnostic accuracy while reducing the workload for physicians.
DINOv2~\cite{oquab2023DINOv2} is a cutting-edge foundation model in computer vision. With a great deal of model parameters, DINOv2 possesses the ability to leverage extensive training data and establish long-term memory, leading to state-of-the-art (SOTA) accuracy across various tasks. While foundation models excel in general tasks, their performance drops in specialized domains like medicine~\cite{cui2024surgical, cui2024endodac}. This is because training entirely new medical foundation models requires vast amounts of labeled medical data and significant computational power, both of which are scarce resources. Nonetheless, pre-training on large-scale general domain vision can effectively initialize our model for fine-tuning downstream tasks. Hence, we introduce our solution to leverage the fine-tuning capabilities of the DINOv2. We aim to maximize the utilization of the pre-trained parameters and harness their benefits for downstream classification tasks in capsule endoscopy diagnosis. In particular, Our contributions are shown as follows:

\begin{itemize}[]
\item We explore the potential of the foundation model DINOv2 for medical image analysis in capsule endoscopy diagnosis problems.

\item We propose a novel fine-tuning and adaptation strategy for DINOv2, utilizing the LoRA Adaptation technique. This approach minimizes additional training costs while effectively tailoring the model to the unique challenges of capsule endoscopy diagnosis.

\item Our method is evaluated on two publicly available datasets. The results demonstrate superior performance compared to other SOTA methods in capsule endoscopy diagnosis. 

\end{itemize}
\section{RELATED WORK}
\subsection{Foundation Models for Images classification}
Convolutional neural networks (CNNs), like ResNet ~\cite{he2016deep}, RegNet~\cite{radosavovic2020designing}, ConvNeXt~\cite{liu2022convnet}, have been sufficiently explored for the classification of both natural and medical images In the past few years. Introduced by Dosovitskiy et al.~\cite{dosovitskiy2020image}, transformers revolutionized sequence modeling with their ability to capture long-range dependencies, which is suitable for tasks that require understanding contextual relationships. However, Vision Transformers (ViT) typically rely on large datasets for pretraining. To overcome this challenge, several approaches have been proposed: Swin Transformer~\cite{liu2021swin} adopts a hierarchical structure by merging image patches and computes attention in non-overlapping local windows, which improves the scalability of ViT and allows for capturing dependencies at different scales. ViTMobile~\cite{mehta2021mobilevit} combines the advantages of both CNNs and ViT. It introduces the MobileViTBlock, which inserts a TransformerBlock between two convolutional layers to learn global representations with an inductive bias. Data-efficient Image Transformer (DeiT)~\cite{touvron2021training} proposed a teacher-student framework based on ViT, which significantly reduces computational resource consumption and training time. Class-Attention in Image Transformer (CaiT)~\cite{touvron2021going} proposes the use of Layer Scale to facilitate the convergence of deep ViT models, which improves efficiency and performance. Moreover, the Mamba framework has gained interest in the context of state space models (SSMs). Mamba incorporates time-varying parameters into SSMs and proposes a hardware-aware algorithm for efficient training and inference. Vision Mamba~\cite{zhu2024vision} specifically focuses on vision tasks and serves as a vision backbone network based on pure SSMs, achieving high performance while significantly improving computational and memory efficiency.

\subsection{Deep Learning for capsule endoscopy diagnosis}
Deep learning algorithms have been extensively explored for disease diagnosis in capsule endoscopy images. To solve the challenges associated with limited data, imbalanced classes, and the lack of distinguishable features in the wireless capsule endoscopy (WCE) dataset, several notable approaches have been proposed. Gjestang et al.~\cite{gjestang2021self} proposed a new network for disease classification using a semi-supervised teacher-student network. This approach leverages unlabeled data to improve itself, achieving better results than traditional supervised learning algorithms. To overcome the need for extensive training samples, Khadka et al. ~\cite{khadka2022meta} introduced the implicit model-agnostic meta-learning (iMAML) method. This method demonstrated high accuracy on unseen datasets with only a small amount of training samples.  However, although the progress made by these methods, the difficulties of imbalanced classes, distinguishable features, and limited data persist in the WCE dataset. To overcome the reliance of ViT on large datasets for pretraining, several approaches have been proposed. Muruganantham et al. ~\cite{muruganantham2022attention} leveraged the potentiality of self-attention mechanisms in computer vision and introduced a dual-branch CNN model to improve accuracy in capsule endoscopy diagnosis. Convolutional vision Transformer (CvT)~\cite{wu2021cvt} aimed to combine the low-level information extraction ability of CNNs with ViT by using Convolutional Token Embedding. Yuan et al.~\cite{yuan2021tokens} introduced Tokens-to-Token (T2T), a method that utilizes a layer-by-layer transformation approach.

Despite progress in training ViT from scratch on natural image datasets, its application in processing WCE data is not yet widespread. Due to the limitations of the WCE dataset, including its small size and high-class imbalance, directly applying existing methods designed for larger datasets proves difficult. These challenges act as a hurdle for the development and widespread use of Vision Transformer (ViT) models on WCE data.

\section{Network Architecture}

\subsection{Network Backbone}
In recent years, foundation models have become increasingly popular. Known for their large parameter sizes, foundation models have shown remarkable performance on various tasks by leveraging long-term memory captured from extensive training data. Introduced by Oquab et al.~\cite{oquab2023DINOv2}, DINOv2 generates vision features at both the image and pixel levels, which can be utilized without any task-specific limitations. To leverage the capabilities of foundation models for capsule endoscopy diagnosis, we propose a method that incorporates low-rank adaptation learning. Our method introduces LoRA layers that adapt to the domain knowledge specific to capsule endoscopy diagnosis. During training, we get the DINO backbone frozen, which produces detailed and accurate visual representations and focuses on optimizing the LoRA layers and classifier. The architecture of our framework is derived from DINOv2. DINOv2 divides images into patches and uses linear projection to flatten them. To incorporate spatial information, positional embeddings are added to the tokens. Additionally, a learnable class token aggregates global image information. Next, the encoded image embeddings are fed through a sequence of Transformer blocks, which refine and update the descriptions of the image content. The core image encoder DINOv2 remains frozen while training. Instead, the focus shifts to adding LoRA layers to each Transformer block. Think of these LoRA layers as assistants that help the main Transformer learn more efficiently. These LoRA layers act like data compressors, squeezing the complex information processed by the Transformer into a more manageable format. Then, they cleverly expand this compressed data back to its original size, ensuring compatibility with the existing Transformer blocks. Each LoRA layer operates independently. This approach allows for more tailored learning within each Transformer block. The incorporation of LoRA layers in our framework allows for fine-tuning and adaptation of foundation models specifically for capsule endoscopy diagnosis. By leveraging low-rank adaptation learning, our method optimizes the LoRA layers while keeping the powerful visual representation capabilities of the frozen DINO image encoder. As shown in Fig 1, our proposed method architecture inherits from DINOv2. We supervise the model's training using cross-entropy loss with label smoothing. We set the label smooth smoothing to 0.1, and choose the Adam optimizer with the learning rate setting to 0.0001.

\begin{figure}
\centering
\includegraphics[scale=0.5]{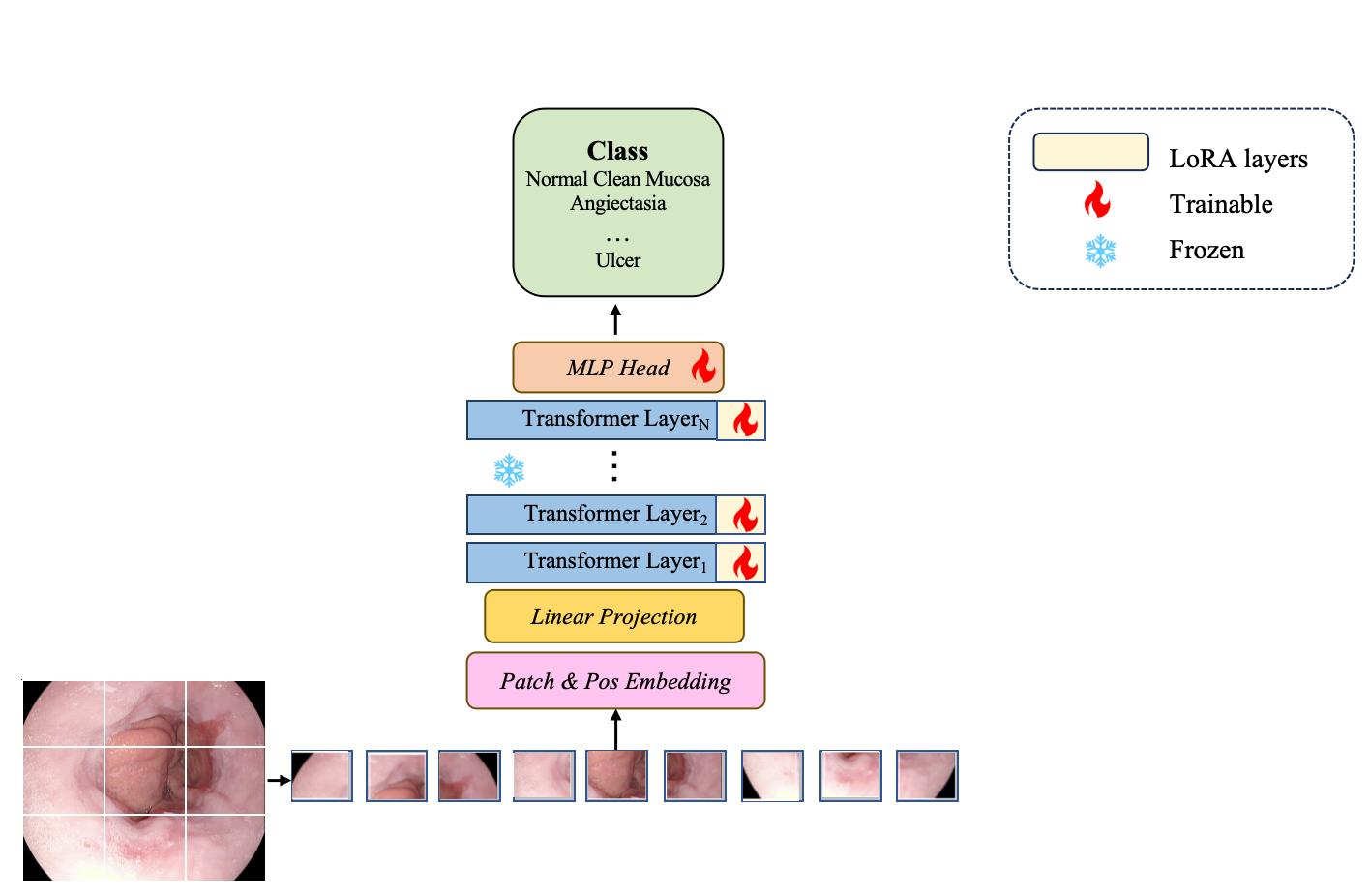}
\caption{Diagram illustrating our proposed network. The input image undergoes tokenization by extracting scaled-down patches, which are then projected linearly. To enhance the embedding ability, both a patch-independent class token and a positional embedding are incorporated. The image encoder remains frozen, while trainable LoRA layers are added for fine-tuning the model.} \label{fig1}
\end{figure}

\begin{figure}
\centering
\includegraphics[scale=0.28]{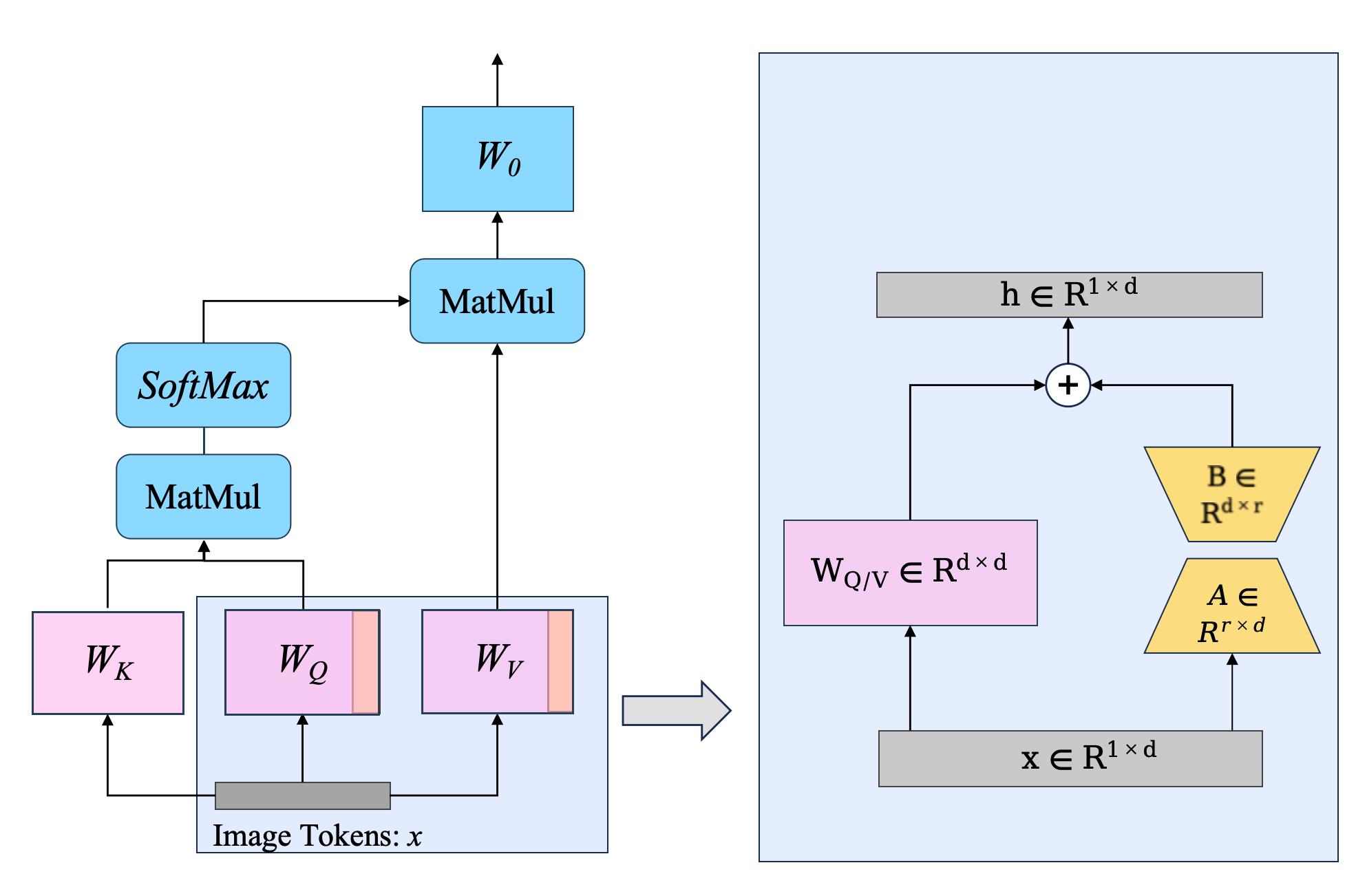}
\caption{The LoRA layers in our framework. LoRA layers are selectively applied to the $q$ and $v$ projection layers within each transformer block. The projection layers corresponding to $q, k, v$ and $o$ are denoted as $W_q, W_k, W_v$ and $W_o$, respectively.} \label{fig1}
\end{figure}
\subsection{LoRA Layers}
LoRA ~\cite{hu2021lora} proposes a method for fine-tuning large language models for specific tasks. It leverages the low-dimensional nature of these pre-trained models, allowing for efficient adaptation with minimal changes to the original parameters. The fundamental idea behind LoRA is that the pre-trained large model's ability to learn effectively is not significantly affected when randomly projected into a smaller subspace. LoRA integrates trainable rank decomposition matrices into every Transformer block while maintaining the frozen pre-trained model weights. This approach results in a substantial reduction in the number of trainable parameters, specifically through the introduction of these trainable rank decomposition matrices for downstream tasks. This reduction in parameters can be beneficial, particularly when applying the model to tasks with limited training data or resource constraints. LoRA enables the fine-tuning of large foundation models with a smaller number of parameters, allowing them to be adapted more efficiently to specific downstream tasks. This technique provides a practical approach to strike a balance between the computational demands of large models and the specific requirements of downstream applications.

Compared to getting the entire model fin-tuned, freezing the model and incorporating trainable LoRA layers lead to substantial reductions in the computational resources required for training. Additionally, this approach facilitates convenient model deployment. The LoRA layers are presented in Fig 2. We followed~\cite{zhang2023customized}, selectively applying low-rank approximation to the $q$ and $v$ projection layers, thereby avoiding excessive impact on attention scores. 

\begin{equation}
Q=\hat{W}_q a=W_q a+B_q A_q a; \;\;
K=W_k a; \;\;
V=\hat{W}_v a=W_v a+B_v A_v a
\end{equation}

With the fundamental formulation of LoRA as described earlier, consider an encoded token embedding $x$. During the multi-head self-attention operation, the processing of $q$, $k$, and $v$ projection layers is modified. Specifically, we have frozen projection layers denoted by $W_q, W_k$, and $W_v$ for $q$, $k$, and $v$, respectively. Additionally, there are trainable LoRA layers represented by $A_q, B_q, A_v$, and $B_v$. The self-attention mechanism remains unchanged, following the description provided by:

\begin{equation}
{Att}(Q,K,V)={Softmax}\left(\frac{Q K^T}{\sqrt{C_{{out}}}}+B\right) V
\end{equation}

where $C_{{out}}$ denotes the numbers of output tokens.

\section{EXPERIMENTS}
\subsection{Datasets}
We conducted experiments on two datasets, the Kvasir-Capsule ~\cite{smedsrud2021kvasir} dataset and the Kvasir-v2 ~\cite{pogorelov2017kvasir} dataset.

\noindent
\textbf{Kvasir-Capsule}: To focus on disease classification, we keep the "Luminal findings" category and remove the three classes related to anatomy from the original dataset. This results in a new dataset with 11 classes: 10 disease categories (Erythema, Ulcer, Blood-hematin, Angiectasia, Pylorus, Reduced Mucosal View, Body, Blood-fresh, Lymphangiectasia, Erosion) and a class for "Normal Clean Mucosa".

\noindent
\textbf{Kvasir-v2}: The Kvasir-v2 dataset including 8 different categories of diseases (dyed-lifted-polyps, esophagitis, normal-z-line, ulcerative-colitis, normal-pylorus, dyed-resection-margins, polyps, normal-cecum). 

We split training and testing sets in a 4:1 ratio.

\subsection{Implementation Details}
We developed the neural networks in Python using the PyTorch framework. We conducted training and validation experiments on a Ubuntu 18.04 LTS server equipped with 6 NVIDIA GeForce RTX 3090 GPUs. Our training configuration included all models with 60 training epochs and a batch size of 64.  We categorized the algorithms into CNNs and ViTs as follows for comparison:
ResNet~\cite{he2016deep}, RegNet~\cite{radosavovic2020designing}, ConvNeXt~\cite{liu2022convnet}; ViT~\cite{dosovitskiy2020image}, Swin-Transformer~\cite{liu2021swin}, DeiT~\cite{touvron2021training}, CaiT~\cite{touvron2021going}, ViTMobile~\cite{mehta2021mobilevit}, VisionMamba~\cite{zhu2024vision}.

\subsection{Results}
We compare our proposed model against SOTA approaches, including three CNN-based methods and six ViT-based methods. The results in Tables 1 and 2 demonstrate the effectiveness of various algorithms for disease classification. Firstly, CNNs achieved impressive performance across both datasets. ConvNeXt, in particular, stood out as the most accurate CNN model, reaching a 95.19\% accuracy on the Kvasir-Capsule dataset and an accuracy of 98.38\% on the Kvasir-v2 dataset. Among ViT algorithms, ViTMobile surpassed the performance of CNNs on the Kvasir-Capsule dataset by 96.09\% and 98.25\% respectively. CaiT, another ViT model, excelled on the Kvasir-v2 dataset, achieving better accuracy than CNNs by 98.75\%. Lastly, Our proposed method, DINOv2-b with LoRA, achieved the overall best performance on both datasets. It reached an accuracy of 97.75\% on the Kvasir-Capsule dataset and an even higher accuracy of 98.81\% on the Kvasir-v2 dataset. Meanwhile, Our method has fewer training parameters and shorter training time.

\begin{table}[h]
\centering
\caption{The results of quantitative comparisons with the SOTA methods on the public Kvasir-Capsule dataset.}
\resizebox{0.8\textwidth}{!}{
\begin{tabular*}{\hsize}{@{\extracolsep{\fill}}cccccc@{}}
\toprule
 Kvasir-Capsule & Acc ({\it{\%}}) & Recall ({\it{\%}})& Precision ({\it{\%}})& F1 Score ({\it{\%}})& Sensitivity ({\it{\%}})\\
\colrule
ResNet~\cite{he2016deep}      & 94.80 &  74.65 & 84.82 &78.83 &74.65  \\
RegNet~\cite{radosavovic2020designing}  & 94.59 &  82.01 & 83.90 &80.41 &82.01  \\
ConvNeXt~\cite{liu2022convnet} & 95.19 &  79.16 & 81.99 &79.06 &79.16  \\
ViT~\cite{dosovitskiy2020image}  & 92.37 &  66.37 & 73.95 &69.09 &66.37  \\
Swin-Transformer~\cite{liu2021swin}& 93.79 &  76.68  & 86.55 &79.95 &76.68  \\
DeiT~\cite{touvron2021training}& 93.58 &  77.89 & 88.95 &80.24 &77.89  \\
CaiT~\cite{touvron2021going} & 94.15 &  83.19 & 81.82 &78.93 &83.91  \\
ViTMobile~\cite{mehta2021mobilevit}& 96.09 &  73.91  & 90.05 &81.92 &73.91 \\
VisionMamba~\cite{zhu2024vision} & 94.97 &  75.34  & 83.20 &76.61 &75.34  \\
\botrule
DINOv2-s~\cite{oquab2023DINOv2}                & 93.28 & 76.64  &74.19  &73.95 &76.64  \\
DINOv2-b~\cite{oquab2023DINOv2}                & 94.98 & 76.89  &78.93  &76.35 &76.89  \\
DINOv2-s with LoRA   &95.11  &73.40   &80.01  &74.34 &73.40  \\
DINOv2-b with LoRA & \textbf{97.75} & \textbf{97.75}  & \textbf{97.83}  & \textbf{87.76} & \textbf{97.75}  \\
\botrule
\end{tabular*}}
\end{table}

\begin{table}[h]
\caption{The results of quantitative comparisons with the SOTA methods on the public Kvasir-v2 dataset.}
\centering
\resizebox{0.8\textwidth}{!}{
\begin{tabular*}{\hsize}{@{\extracolsep{\fill}}cccccc@{}}
\toprule
Kvasir-v2 & Acc ({\it{\%}}) & Recall ({\it{\%}})& Precision ({\it{\%}})& F1 Score ({\it{\%}})& Sensitivity ({\it{\%}})\\
\colrule
ResNet~\cite{he2016deep}  & 96.06 & 96.06 & 96.14 & 96.05 &96.06\\
RegNet~\cite{radosavovic2020designing}& 97.88 & 97.88 &97.89  & 97.88 &97.88\\
ConvNeXt~\cite{liu2022convnet} & 98.38 & 97.88 &97.89  & 97.88 &97.88\\
ViT~\cite{dosovitskiy2020image} & 82.56 & 82.56 &82.89  &82.48  &82.56\\
DeiT~\cite{touvron2021training} & 98.25 & 98.25 &98.27  & 98.25 &98.25\\
CaiT~\cite{touvron2021going}  & 98.75 & 98.75 &98.75  &98.75  &98.75\\
Swin-Transformer~\cite{liu2021swin} & 95.08 & 82.99 &74.48  &75.96  &82.99\\
ViTMobile~\cite{mehta2021mobilevit} & 98.25 & 98.25 &98.26  &98.25  &98.25\\
VisionMamba~\cite{zhu2024vision}& 97.81 & 97.81 &97.84  &97.81  &97.81\\
\botrule
DINOv2-s~\cite{oquab2023DINOv2}  & 88.00 & 89.00  &88.43  &87.85  &88.00\\
DINOv2-b~\cite{oquab2023DINOv2}                        & 89.91 & 89.81 &89.90  &89.74  &89.81\\
DINOv2-s with LoRA           &94.69   &94.69 &94.72   &94.67  &94.69\\
DINOv2-b with LoRA      & \textbf{98.81} & \textbf{98.81} & \textbf{98.81} & \textbf{98.81}  & \textbf{98.81}\\
\botrule
\end{tabular*}}
\end{table}

\section{CONCLUSIONS}
In our work, we adapt the vision foundation model DINOv2 with the LoRA adapter for capsule endoscopy diagnosis. To adapt the DINOv2 model to the surgical domain, we incorporate the LoRA layers. These layers allow for fine-tuning the network with minimal additional parameters, enabling the model to effectively capture the unique characteristics of surgical data. To evaluate the performance of our proposed method, we conducted comparison experiments on two publicly available capsule endoscopy diagnosis datasets. We compared our methodology with several CNN and ViT algorithms commonly used in computer vision tasks. Our approach outperforms all the comparison methods. These results emphasize the potential of vision foundation models such as DINOv2, in the field of surgical domain and contribute to the advancement of automated diagnosis in capsule endoscopy. 

\section*{ACKNOWLEDGEMENT} 
This work was supported by Hong Kong Research Grants Council (RGC) Collaborative Research Fund (CRF) C4026-21GF, General Research Fund (GRF) 14203323, 14216022 \& 14211420, NSFC/RGC Joint Research Scheme N\_CUHK420/22, Shenzhen-Hong Kong-Macau Technology Research Programme (Type C) STIC Grant 202108233000303, and Regional Joint Fund Project 2021B1515120035 (B.02.21.00101) of Guangdong Basic and Applied Research Fund.








\clearpage

\normalMode

\bibliographystyle{plain}
\bibliography{ref}

\begin{thebibliography}{10}

\bibitem{bai2023llcaps}
Long Bai, Tong Chen, Yanan Wu, An~Wang, Mobarakol Islam, and Hongliang Ren.
\newblock Llcaps: Learning to illuminate low-light capsule endoscopy with curved wavelet attention and reverse diffusion.
\newblock In {\em International Conference on Medical Image Computing and Computer-Assisted Intervention}, pages 34--44. Springer, 2023.

\bibitem{bai2022transformer}
Long Bai, Liangyu Wang, Tong Chen, Yuanhao Zhao, and Hongliang Ren.
\newblock Transformer-based disease identification for small-scale imbalanced capsule endoscopy dataset.
\newblock {\em Electronics}, 11(17):2747, 2022.

\bibitem{cui2024surgical}
Beilei Cui, Mobarakol Islam, Long Bai, and Hongliang Ren.
\newblock Surgical-dino: adapter learning of foundation models for depth estimation in endoscopic surgery.
\newblock {\em International Journal of Computer Assisted Radiology and Surgery}, pages 1--8, 2024.

\bibitem{cui2024endodac}
Beilei Cui, Mobarakol Islam, Long Bai, An~Wang, and Hongliang Ren.
\newblock Endodac: Efficient adapting foundation model for self-supervised depth estimation from any endoscopic camera.
\newblock {\em arXiv preprint arXiv:2405.08672}, 2024.

\bibitem{dosovitskiy2020image}
Alexey Dosovitskiy, Lucas Beyer, Alexander Kolesnikov, Dirk Weissenborn, Xiaohua Zhai, Thomas Unterthiner, Mostafa Dehghani, Matthias Minderer, Georg Heigold, Sylvain Gelly, et~al.
\newblock An image is worth 16x16 words: Transformers for image recognition at scale.
\newblock {\em arXiv preprint arXiv:2010.11929}, 2020.

\bibitem{gjestang2021self}
Henrik~L Gjestang, Steven~A Hicks, Vajira Thambawita, P{\aa}l Halvorsen, and Michael~A Riegler.
\newblock A self-learning teacher-student framework for gastrointestinal image classification.
\newblock In {\em 2021 IEEE 34th International Symposium on Computer-Based Medical Systems (CBMS)}, pages 539--544. IEEE, 2021.

\bibitem{he2016deep}
Kaiming He, Xiangyu Zhang, Shaoqing Ren, and Jian Sun.
\newblock Deep residual learning for image recognition.
\newblock In {\em Proceedings of the IEEE conference on computer vision and pattern recognition}, pages 770--778, 2016.

\bibitem{hu2021lora}
Edward~J Hu, Yelong Shen, Phillip Wallis, Zeyuan Allen-Zhu, Yuanzhi Li, Shean Wang, Lu~Wang, and Weizhu Chen.
\newblock Lora: Low-rank adaptation of large language models.
\newblock {\em arXiv preprint arXiv:2106.09685}, 2021.

\bibitem{khadka2022meta}
Rabindra Khadka, Debesh Jha, Steven Hicks, Vajira Thambawita, Michael~A Riegler, Sharib Ali, and P{\aa}l Halvorsen.
\newblock Meta-learning with implicit gradients in a few-shot setting for medical image segmentation.
\newblock {\em Computers in Biology and Medicine}, 143:105227, 2022.

\bibitem{li2023semi}
Hechen Li, Yanan Wu, Long Bai, An~Wang, Tong Chen, and Hongliang Ren.
\newblock Semi-supervised learning for segmentation of bleeding regions in video capsule endoscopy.
\newblock {\em Procedia Computer Science}, 226:29--35, 2023.

\bibitem{liu2021swin}
Ze~Liu, Yutong Lin, Yue Cao, Han Hu, Yixuan Wei, Zheng Zhang, Stephen Lin, and Baining Guo.
\newblock Swin transformer: Hierarchical vision transformer using shifted windows.
\newblock In {\em Proceedings of the IEEE/CVF international conference on computer vision}, pages 10012--10022, 2021.

\bibitem{liu2022convnet}
Zhuang Liu, Hanzi Mao, Chao-Yuan Wu, Christoph Feichtenhofer, Trevor Darrell, and Saining Xie.
\newblock A convnet for the 2020s.
\newblock In {\em Proceedings of the IEEE/CVF conference on computer vision and pattern recognition}, pages 11976--11986, 2022.

\bibitem{mehta2021mobilevit}
Sachin Mehta and Mohammad Rastegari.
\newblock Mobilevit: light-weight, general-purpose, and mobile-friendly vision transformer.
\newblock {\em arXiv preprint arXiv:2110.02178}, 2021.

\bibitem{muruganantham2022attention}
Prabhananthakumar Muruganantham and Senthil~Murugan Balakrishnan.
\newblock Attention aware deep learning model for wireless capsule endoscopy lesion classification and localization.
\newblock {\em Journal of Medical and Biological Engineering}, 42(2):157--168, 2022.

\bibitem{oquab2023DINOv2}
Maxime Oquab, Timoth{\'e}e Darcet, Th{\'e}o Moutakanni, Huy Vo, Marc Szafraniec, Vasil Khalidov, Pierre Fernandez, Daniel Haziza, Francisco Massa, Alaaeldin El-Nouby, et~al.
\newblock Dinov2: Learning robust visual features without supervision.
\newblock {\em arXiv preprint arXiv:2304.07193}, 2023.

\bibitem{pogorelov2017kvasir}
Konstantin Pogorelov, Kristin~Ranheim Randel, Carsten Griwodz, Sigrun~Losada Eskeland, Thomas de~Lange, Dag Johansen, Concetto Spampinato, Duc-Tien Dang-Nguyen, Mathias Lux, Peter~Thelin Schmidt, et~al.
\newblock Kvasir: A multi-class image dataset for computer aided gastrointestinal disease detection.
\newblock In {\em Proceedings of the 8th ACM on Multimedia Systems Conference}, pages 164--169, 2017.

\bibitem{radosavovic2020designing}
Ilija Radosavovic, Raj~Prateek Kosaraju, Ross Girshick, Kaiming He, and Piotr Doll{\'a}r.
\newblock Designing network design spaces.
\newblock In {\em Proceedings of the IEEE/CVF conference on computer vision and pattern recognition}, pages 10428--10436, 2020.

\bibitem{smedsrud2021kvasir}
Pia~H Smedsrud, Vajira Thambawita, Steven~A Hicks, Henrik Gjestang, Oda~Olsen Nedrejord, Espen N{\ae}ss, Hanna Borgli, Debesh Jha, Tor Jan~Derek Berstad, Sigrun~L Eskeland, et~al.
\newblock Kvasir-capsule, a video capsule endoscopy dataset.
\newblock {\em Scientific Data}, 8(1):142, 2021.

\bibitem{tan2024endoood}
Qiaozhi Tan, Long Bai, Guankun Wang, Mobarakol Islam, and Hongliang Ren.
\newblock Endoood: Uncertainty-aware out-of-distribution detection in capsule endoscopy diagnosis.
\newblock {\em arXiv preprint arXiv:2402.11476}, 2024.

\bibitem{touvron2021training}
Hugo Touvron, Matthieu Cord, Matthijs Douze, Francisco Massa, Alexandre Sablayrolles, and Herv{\'e} J{\'e}gou.
\newblock Training data-efficient image transformers \& distillation through attention.
\newblock In {\em International conference on machine learning}, pages 10347--10357. PMLR, 2021.

\bibitem{touvron2021going}
Hugo Touvron, Matthieu Cord, Alexandre Sablayrolles, Gabriel Synnaeve, and Herv{\'e} J{\'e}gou.
\newblock Going deeper with image transformers.
\newblock In {\em Proceedings of the IEEE/CVF international conference on computer vision}, pages 32--42, 2021.

\bibitem{wang2023rethinking}
Guankun Wang, Long Bai, Yanan Wu, Tong Chen, and Hongliang Ren.
\newblock Rethinking exemplars for continual semantic segmentation in endoscopy scenes: Entropy-based mini-batch pseudo-replay.
\newblock {\em Computers in Biology and Medicine}, 165:107412, 2023.

\bibitem{wu2021cvt}
Haiping Wu, Bin Xiao, Noel Codella, Mengchen Liu, Xiyang Dai, Lu~Yuan, and Lei Zhang.
\newblock Cvt: Introducing convolutions to vision transformers.
\newblock In {\em Proceedings of the IEEE/CVF international conference on computer vision}, pages 22--31, 2021.

\bibitem{yuan2021tokens}
Li~Yuan, Yunpeng Chen, Tao Wang, Weihao Yu, Yujun Shi, Zi-Hang Jiang, Francis~EH Tay, Jiashi Feng, and Shuicheng Yan.
\newblock Tokens-to-token vit: Training vision transformers from scratch on imagenet.
\newblock In {\em Proceedings of the IEEE/CVF international conference on computer vision}, pages 558--567, 2021.

\bibitem{zhang2023customized}
Kaidong Zhang and Dong Liu.
\newblock Customized segment anything model for medical image segmentation.
\newblock {\em arXiv preprint arXiv:2304.13785}, 2023.

\bibitem{zhang2022deep}
Yameng Zhang, Long Bai, Li~Liu, Hongliang Ren, and Max Q-H Meng.
\newblock Deep reinforcement learning-based control for stomach coverage scanning of wireless capsule endoscopy.
\newblock In {\em 2022 IEEE International Conference on Robotics and Biomimetics (ROBIO)}, pages 01--06. IEEE, 2022.

\bibitem{zhu2024vision}
Lianghui Zhu, Bencheng Liao, Qian Zhang, Xinlong Wang, Wenyu Liu, and Xinggang Wang.
\newblock Vision mamba: Efficient visual representation learning with bidirectional state space model.
\newblock {\em arXiv preprint arXiv:2401.09417}, 2024.

\end{thebibliography}
 
\end{document}